\documentclass[runningheads]{llncs}
\usepackage{graphicx}
\newcommand{\LL}{\ensuremath{\mathbf{L}}}

\begin{document}

%more space for tables
\renewcommand{\arraystretch}{1.4}
\setlength{\tabcolsep}{0.3em}

\title{3D MRI brain tumor segmentation using autoencoder regularization}
\titlerunning{3D MRI brain tumor segmentation using autoencoder regularization}
\author{Andriy Myronenko}
\authorrunning{A. Myronenko}
\institute{NVIDIA, Santa Clara, CA \\ \email{amyronenko@nvidia.com} }
\maketitle              % typeset the header of the contribution
\begin{abstract}

Automated segmentation of brain tumors from 3D magnetic resonance images (MRIs) is necessary for the diagnosis, monitoring, and treatment planning of the disease. Manual delineation practices require anatomical knowledge, are expensive, time consuming and can be inaccurate due to human error. Here, we describe a semantic segmentation network for tumor subregion segmentation from 3D MRIs based on encoder-decoder architecture.  Due to a limited training dataset size, a variational auto-encoder branch is added to reconstruct the input image itself in order to regularize the shared decoder and impose additional constraints on its layers. The current approach won 1st place in the BraTS 2018 challenge. 

\end{abstract}

\section{Introduction}

Brain tumors are categorized into primary and secondary tumor types.  Primary brain tumors 	 originate from brain cells, whereas  secondary tumors metastasize into the brain from other organs. The most common type of primary brain tumors are gliomas, which arise from brain glial cells.  Gliomas can be of low-grade  (LGG) and high-grade (HGG) subtypes. High grade gliomas are an aggressive  type of malignant brain tumor that grow rapidly, usually require surgery and radiotherapy and have poor survival prognosis. Magnetic Resonance Imaging (MRI) is a key diagnostic tool for brain tumor analysis, monitoring and surgery planning. Usually, several complimentary 3D MRI modalities are acquired - such as T1, T1 with contrast agent (T1c), T2 and Fluid Attenuation Inversion Recover (FLAIR) - to emphasize different tissue properties and areas of tumor spread.  For example the contrast agent, usually gadolinium, emphasizes hyperactive tumor subregions in T1c MRI modality.  

Automated segmentation of 3D brain tumors can save physicians time and provide an accurate reproducible solution for further tumor analysis and monitoring. Recently, deep learning based segmentation techniques surpassed traditional computer vision methods for dense semantic segmentation. Convolutional neural networks (CNN)  are able to learn from examples and demonstrate state-of-the-art segmentation accuracy both in 2D natural images~\cite{deeplabv3plus2018} and in 3D medical image modalities~\cite{Milletari16}. 

\begin{figure}[t] 
 	\centering
 	\includegraphics[clip=true, trim=0pt 0pt 0pt 0pt, width=0.9\textwidth]{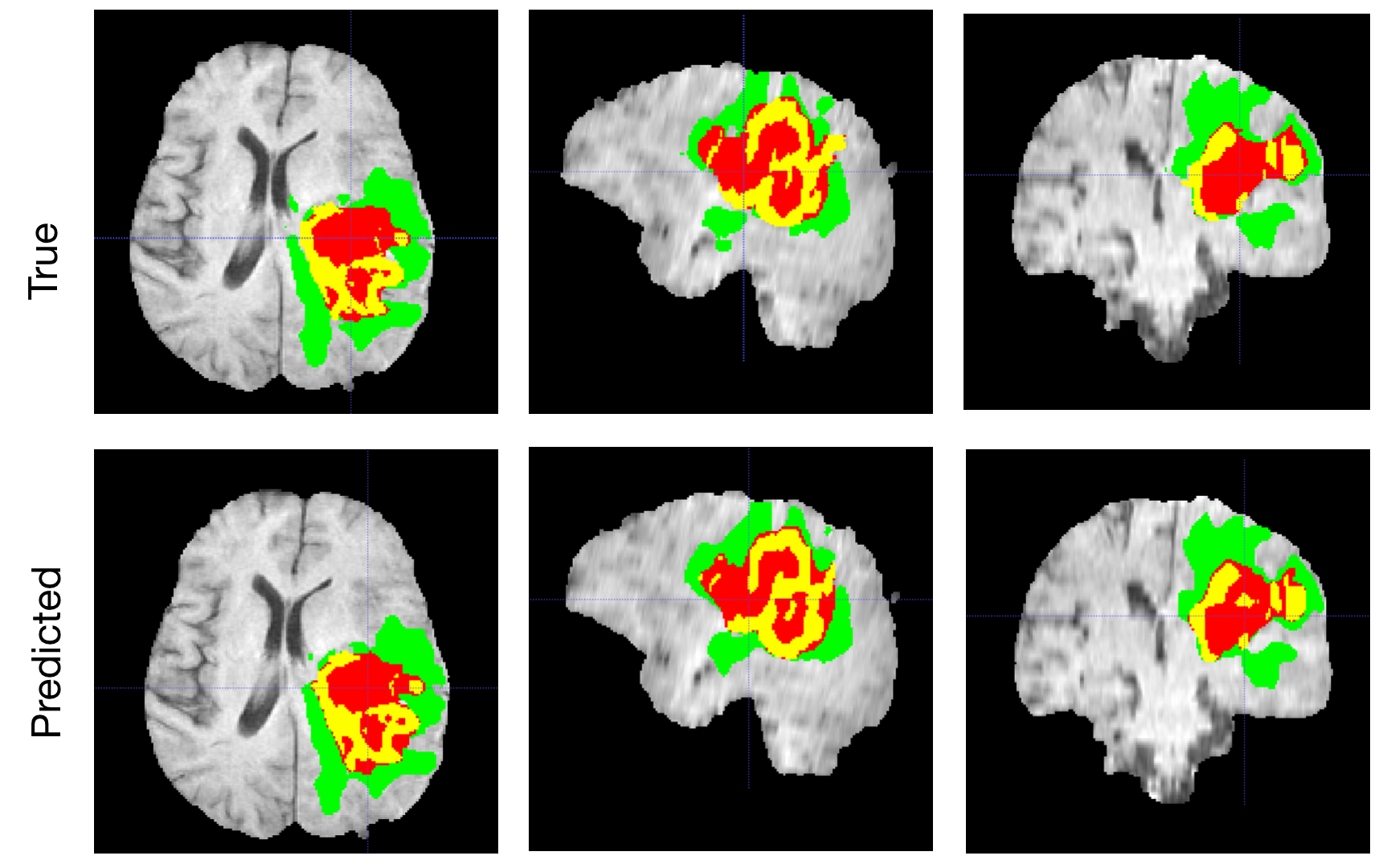}
 	\caption{A typical segmentation example with true and predicted labels overlaid over T1c MRI axial, sagittal and coronal slices.  The whole tumor (WT) class includes all visible labels (a union of green, yellow and red labels), the tumor core (TC) class is a union of red and yellow, and the enhancing tumor core (ET) class is shown in yellow (a hyperactive tumor part). The predicted segmentation results match the ground truth well.}
 	\label{fig:seg}
 	\vspace{-5mm}
 \end{figure}

Multimodal Brain Tumor Segmentation Challenge (BraTS) aims to evaluate state-of-the-art methods for the segmentation of brain tumors by providing a 3D MRI dataset with ground truth tumor segmentation labels annotated by physicians~\cite{BratsAll2018,brats1,brats2,brats3,brats4}. This year, BraTS 2018 training dataset included 285 cases (210 HGG and 75 LGG), each with four 3D MRI modalities (T1, T1c, T2 and FLAIR) rigidly aligned, resampled to 1x1x1 mm isotropic resolution and skull-stripped. The input image size is 240x240x155. The data were collected from 19 institutions, using various MRI scanners. Annotations include 3 tumor subregions: the enhancing tumor, the peritumoral edema, and the necrotic and non-enhancing tumor core.  The annotations were combined into 3 nested subregions: whole tumor (WT), tumor core (TC) and enhancing tumor (ET), as shown in Figure~\ref{fig:seg}. Two additional datasets without the ground truth labels were provided for validation and testing. These datasets required participants to upload the segmentation masks to the organizers' server for evaluations. The validation dataset (66 cases) allowed multiple submissions and was designed for intermediate evaluations. The testing dataset (191 cases) allowed only a single submission, and was used to calculate the final challenge ranking.  
%This year, BraTS 2018 attracted more than 390 participating teams.  

In this work, we describe our semantic segmentation approach for volumetric 3D brain tumor segmentation from multimodal 3D MRIs, which won the BraTS 2018 challenge. We follow the encoder-decoder structure of CNN, with asymmetrically large encoder to extract deep image features, and the decoder part reconstructs dense segmentation masks. We also add the variational autoencoder (VAE) branch to the network to reconstruct the input images jointly with segmentation in order to regularize the shared encoder. At inference time, only the main segmentation encode-decoder part is used.

\section{Related work}
\label{sec:relatedwork}

Last year, BraTS 2017, top performing submissions included Kamnitsas et al.~\cite{Kamnitsas17} who proposed to ensemble several models for robust segmentation (EMMA), and  Wang et al.~\cite{Wang17} who proposed to segment tumor subregions in cascade using anisotropic convolutions. EMMA takes advantage of an ensemble of several independently trained architectures. In particular, EMMA combined DeepMedic~\cite{Kamnitsas16}, FCN~\cite{Long15} and U-net~\cite{Ronneberger15} models and ensembled their segmentation predictions. During training they used a batch size of 8, and a crop of 64x64x64 3D patch. EMMA's ensemble of different models demonstrated a good generalization performance winning the BraTS 2017 challenge. Wang et al.~\cite{Wang17} second place paper took a different approach, by training 3 networks for each tumor subregion in cascade, with each subsequent network taking the output of the previous network (cropped) as its input.  Each network was similar in structure and consists of a large encoder part (with dilated convolutions) and a basic decoder. They also decompose the 3x3x3 convolution kernel into intra-slice (3x3x1) and inter-slice (1x1x3) kernel to save on both the GPU memory and the computational time. 

This year, BraTS 2018 top performing submission (in addition to the current work) included Isensee et al.~\cite{Isensee18brats} in the 2nd place, McKinly et al.~\cite{McKinley18brats} and Zhou et al.~\cite{Zhou18brats}, who shared the 3rd place.  Isensee et al.~\cite{Isensee18brats} demonstrated that a generic U-net architecture with a few  minor modifications is enough to achieve competitive performance. 
%The modifications included leaky ReLU, reducing the number of feature maps before upsampling and Instance Normalization~\cite{Ulyanov16} (instead of BatchNorm~\cite{Ioffe15}).
 The authors used a batch size of 2 and a crop size of 128x128x128. Furthermore, the authors used an additional training data from their own institution (which yielded some improvements for the enhancing tumor dice).
 %and substantial  data augmentation(random rotations, flips, scaling,  elastic deformations, gamma correction).
 McKinly et al.~\cite{McKinley18brats} proposed a segmentation CNN in which a DenseNet~\cite{huang2017densely} structure with dilated convolutions was embedded in U-net-like network.  The authors also introduce a new loss function, a generalization of binary cross-entropy, to account for label uncertainty. Finally, Zhou et al.~\cite{Zhou18brats} proposed to use an ensemble of different networks: taking into account multi-scale context information,  segmenting 3 tumor subregions in cascade with a shared backbone weights and adding an attention block.  

Compared to the related works,  we use the largest crop size of 160x192x128 but compromise the batch size to be 1 to be able to fit network into the GPU memory limits. We also output all 3 nested tumor subregions  directly after the sigmoid (instead of using several networks or the softmax over the number of classes). Finally, we add an additional branch to regularize the shared encoder, used only during training. We did not use any additional training data and used only the provided training set. 

 \begin{figure}[h] 
	\centering
	\includegraphics[clip=true, trim=0pt 0pt 0pt 0pt, width=0.9\textwidth]{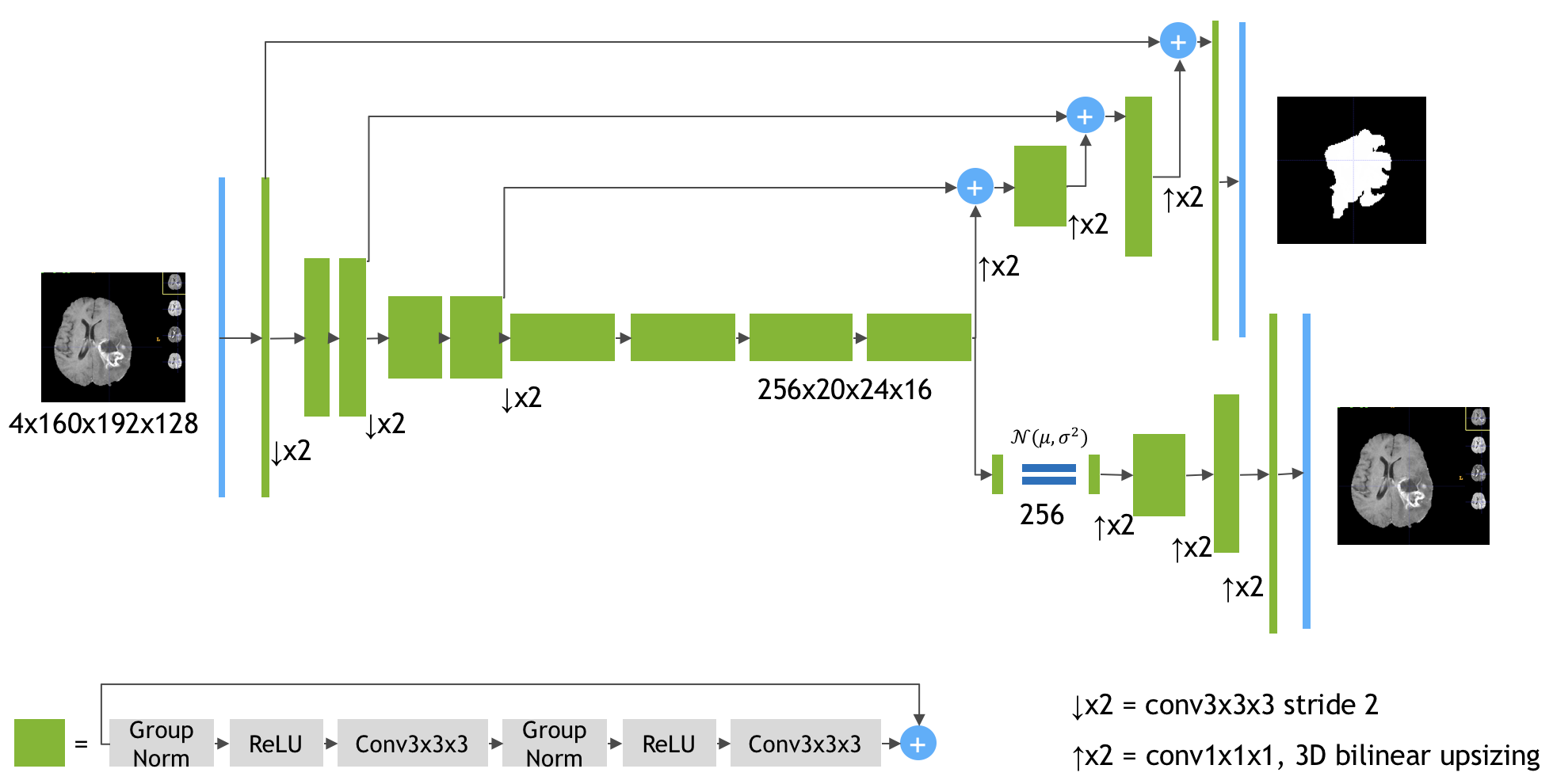}
	\caption{Schematic visualization of the network architecture.  Input is a four channel 3D MRI crop, followed by initial 3x3x3 3D convolution with 32 filters. Each green block is a ResNet-like block with the GroupNorm normalization. The output of the segmentation decoder has three channels (with the same spatial size as the input) followed by a sigmoid for segmentation maps of the three tumor subregions (WT, TC, ET).  The VAE branch reconstructs the input image into itself, and is used only during training to regularize the shared encoder.}
	\label{fig:network}
	\vspace{-5mm}
\end{figure}

\section{Methods}
\label{sec:methods}

Our segmentation approach follows encoder-decoder based CNN architecture with an asymmetrically larger encoder to extract image features and a smaller decoder to reconstruct the segmentation mask~\cite{deeplabv3plus2018,deeplabv3plus2017,he2017maskrcnn,Ronneberger15,Milletari16}.  We add an additional branch to the encoder endpoint to reconstruct the original image, similar to auto-encoder architecture.  The motivation for using the auto-encoder branch is to  add additional guidance and regularization to the encoder part, since the training dataset size is limited.  We follow the variational auto-encoder (VAE) approach to better cluster/group the features of the encoder endpoint. We  describe the building parts of our networks in the next subsections (see also Figure~\ref{fig:network})

\subsection{Encoder part}
The encoder part uses  ResNet~\cite{He16}  blocks, where each block consists of two convolutions with normalization and ReLU, followed by additive identity skip connection. For normalization, we use Group Normalization (GN)~\cite{Wu18}, which shows better than BatchNorm performance when batch size is small (bath size of 1 in our case).  We follow a common CNN approach to progressively downsize image dimensions by 2 and simultaneously increase feature size by 2.  For downsizing we use strided convolutions.   All convolutions are 3x3x3 with initial number of filters equal to 32.  
%The encoder part structure is shown in Table~\ref{tab:encoder}. 
The encoder endpoint has size 256x20x24x16, and is 8 times spatially smaller than the input image. We decided against further downsizing to preserve more spatial content. 

%\begin{table}
%	\centering
%	%\small
%	\caption{Encoder structure, where GN stands for group normalization (with group size of 8), Conv - 3x3x3 convolution, AddId - addition of identity/skip connection. Repeat column shows the number of repetitions of the block. We refer to the final output of the encoder, as the encoder endpoint}
%	\label{tab:encoder}
%	\begin{tabular}{|l|c|c|c|}
%		 \hline
%		Name & Ops & Repeat &Output size    \\ \hline
%		Input & &  &4x160x192x128    \\
%		InitConv & Conv & 1 & 32x160x192x128    \\
%		EncoderBlock0 & GN,ReLU,Conv,GN,ReLU,Conv, AddId & 1 &32x160x192x128    \\
%		EncoderDown1 & Conv stride 2 & 1&64x80x96x64    \\
%		EncoderBlock1 & GN,ReLU,Conv,GN,ReLU,Conv, AddId& 2 &64x80x96x64    \\
%		EncoderDown2 & Conv stride 2& 1&128x40x48x32    \\
%		EncoderBlock2 & GN,ReLU,Conv,GN,ReLU,Conv, AddId& 2 &128x40x48x32    \\
%		EncoderDown3 & Conv stride 2& 1 &256x20x24x16    \\
%		EncoderBlock3 & GN,ReLU,Conv,GN,ReLU,Conv, AddId& 4&256x20x24x16   \\
%		\hline
%	\end{tabular}
%\end{table}

 \subsection{Decoder part}

The decoder structure is similar to the encoder one, but with a single block per each spatial level. Each decoder level begins with upsizing: reducing the number of features  by a factor of 2 (using 1x1x1 convolutions) and doubling the spatial dimension (using 3D bilinear upsampling),  followed by an addition of encoder output of the equivalent spatial level. The end of the decoder has the same spatial size as the original image, and the number of features equal to the initial input feature size, followed by 1x1x1 convolution into 3 channels and a sigmoid function. 

%The decoder structure is shown in Table~\ref{tab:decoder}.
%
%\begin{table}
%	\centering
%	%\small
%	\caption{Decoder structure, where GN stands for group normalization (with group size of 8), Conv - 3x3x3 convolution, Conv1 - 1x1x1 convolution, AddId - addition of identity/skip connection, UpLinear - 3D linear spatial upsampling }
%	\label{tab:decoder}
%	\begin{tabular}{|l|c|c|c|}
%		\hline
%		Name & Ops & Repeat &Output size    \\ \hline
%		DecoderUp2 & Conv1, UpLinear, +EncoderBlock2 &  1 &128x40x48x32    \\
%		DecoderBlock2 & GN,ReLU,Conv,GN,ReLU,Conv, AddId & 1 & 128x40x48x32    \\
%		DecoderUp1 & Conv1, UpLinear, +EncoderBlock1 &  1 &64x80x96x64    \\
%		DecoderBlock1 & GN,ReLU,Conv,GN,ReLU,Conv, AddId & 1 & 64x80x96x64   \\
%		DecoderUp0 & Conv1, UpLinear, +EncoderBlock0 &  1 &32x160x192x128    \\
%		DecoderBlock0 & GN,ReLU,Conv,GN,ReLU,Conv, AddId & 1 & 32x160x192x128   \\
%		DecoderEnd & Conv1, Sigmoid &  1 &1x160x192x144    \\
%		\hline
%	\end{tabular}
%\end{table}

 \subsection{VAE part}
 Starting from the encoder endpoint output, we first reduce the input to a low dimensional space of 256  (128 to represent mean, and 128 to represent std). Then, a sample is drawn from the Gaussian distribution with the given mean and std, and reconstructed into the input image dimensions following the same architecture as the decoder, except we don't use the inter-level skip connections from the encoder here. The VAE part structure is shown in Table~\ref{tab:vaebranch}.
 
 \begin{table}
 	\centering
 	%\small
 	\caption{VAE decoder branch structure, where GN stands for group normalization (with group size of 8), Conv - 3x3x3 convolution, Conv1 - 1x1x1 convolution, AddId - addition of identity/skip connection, UpLinear - 3D linear spatial upsampling, Dense - fully connected layer }
 	\label{tab:vaebranch}
 	\begin{tabular}{|l|c|c|c|}
 		\hline
 		Name & Ops & Repeat &Output size    \\ \hline
 		VD &  GN, ReLU, Conv (16) stride 2, Dense (256) & 1 &  256x1 \\
 		VDraw & sample $\sim \mathcal{N}(\mu (128),\,\sigma^{2} (128))$ & 1 & 128x1 \\
 		VU &  Dense, ReLU, Conv1, UpLinear & 1 &  256x20x24x16 \\
		VUp2 & Conv1, UpLinear &  1 &128x40x48x32    \\
		VBlock2 & GN,ReLU,Conv,GN,ReLU,Conv, AddId & 1 & 128x40x48x32   \\
		VUp1 & Conv1, UpLinear, &  1 &64x80x96x64    \\
		VBlock1 & GN,ReLU,Conv,GN,ReLU,Conv, AddId & 1 & 64x80x96x64   \\
		VUp0 & Conv1, UpLinear, &  1 &32x160x192x128    \\
		VBlock0 & GN,ReLU,Conv,GN,ReLU,Conv, AddId & 1 & 32x160x192x128  \\
 		Vend & Conv1 & 1 & 4x160x192x128 \\
 		\hline
 	\end{tabular}
 \end{table}

 \subsection{Loss}
 Our loss function consists of 3 terms:
  \begin{equation}
 \label{eq:loss}
 \LL =\LL_{dice} + 0.1*\LL_{\textsc{L2}} + 0.1*\LL_{\textsc{KL}}   
 \end{equation}
$\LL_{dice}$ is a soft dice loss~\cite{Milletari16} applied to the  decoder output $p_{pred}$ to match the segmentation mask $p_{true}$:
   \begin{equation}
  \label{eq:dice}
  \LL_{dice}  = \frac{2*\sum p_{true} * p_{pred} }{\sum p_{true}^2 + \sum p_{pred}^2 + \epsilon}   
  \end{equation}
where summation is voxel-wise, and $\epsilon$ is a small constant to avoid zero division. Since the output of the segmentation decoder has 3 channels (predictions for each tumor subregion), we simply add the three dice loss functions together. 

$\LL_{\textsc{L2}}$ is an L2 loss on the VAE branch output $I_{pred}$ to match the input image $I_{input}$:
   \begin{equation}
  	\label{eq:vael2}
  	\LL_{\textsc{L2}}   =|| I_{input} - I_{pred}||_2^2
  \end{equation}
  
$\LL_{\textsc{KL}}$  is standard VAE penalty term~\cite{Kingma14,doersch2016vae}, a KL divergence between the estimated normal distribution $\mathcal{N}(\mu,\sigma^2)$ and a prior distribution $\mathcal{N}(0,1)$, which has a closed form representation:
 \begin{equation}
\label{eq:vaekl}
\LL_{\textsc{KL}} =  \frac{1}{N} \sum  \mu^2  + \sigma^2 -  \log \sigma^2 -1
\end{equation}
where N is total number of image voxels. We empirically found a hyper-parameter weight of $0.1$ to provide a good balance between dice and VAE loss terms in Equation~\ref{eq:loss}.

 \subsection{Optimization}
 We use Adam optimizer with initial learning rate of $ \alpha_{0} = 1e-4$ and progressively decrease it according to:
 \begin{equation}
 \label{eq:learningrate}
 \alpha = \alpha_{0} *\left(1-\frac{e}{N_{e}}\right)^{0.9}  
 \end{equation}
 where $e$ is an epoch counter, and $N_{e}$ is a total number of epochs (300 in our case). 
 We use batch size of 1, and draw input images in random order (ensuring that each training image is drawn once per epoch). 
 
 \subsection{Regularization}
 We use L2 norm regularization on the convolutional kernel parameters with a weight of $1e-5$.  We also use the spatial dropout with a rate of $0.2$  after the initial encoder convolution. We have experimented with other placements of the dropout (including placing dropout layer after each convolution), but did not find any additional accuracy improvements. 
 
  \subsection{Data preprocessing and augmentation}
  We normalize all input images to have zero mean and unit std (based on non-zero voxels only). We  apply a random (per channel) intensity  shift ($-0.1..0.1$  of image std) and scale ($0.9..1.1$) on input image channels.  We also apply a random axis mirror flip (for all 3 axes) with a probability $0.5$. 

 \section{Results}
 \label{sec:results}
 
 We implemented our network in Tensorflow~\cite{tensorflow2015} and trained it on NVIDIA Tesla V100 32GB GPU using BraTS 2018 training dataset (285 cases) without any additional in-house data. During training we used a random crop of size 160x192x128, which ensures that most image content remains within the crop area. We concatenated  4 available 3D MRI modalities into the 4 channel image as an input. The output of the network is 3 nested tumor subregions (after the sigmoid).
 
 We report the results of our approach  on BraTS 2018 validation (66 cases) and the testing sets (191 cases). These datasets were provided with unknown glioma grade and unknown segmentation. We uploaded our segmentation results to the BraTS 2018 server for evaluation of per class dice, sensitivity, specificity and Hausdorff distances. 
 
  Aside from evaluating a single model,  we also applied test time augmentation (TTA) by mirror flipping the input 3D image axes, and averaged the output of the resulting 8 flipped segmentation probability maps. Finally, we ensembled a set of 10 models (trained from scratch) to further improve the performance.

 Table~\ref{tab:valid} shows the results of our model on the BraTS 2018 validation dataset. At the time of initial short paper submission (Jul 13, 2018), our dice accuracy performance was second best (team name  NVDLMED\footnote{https://www.cbica.upenn.edu/BraTS18/lboardValidation.html}) for all of the 3 segmentation labels  (ET, WT, TC). 
 The TTA only marginally improved the performance, but the ensemble of 10 models resulted in ~1\% improvement, which is consistent with the literature results of using ensembles. 
 
 For the testing dataset, only a single submission was allowed. Our results are shown in  Table~\ref{tab:testresults}, which won the 1st place at BraTS 2018 challenge.

\begin{table}
	\centering
	%\small
	\caption{BraTS 2018 validation dataset results. Mean Dice and Hausdorff measurements of the proposed segmentation method. EN - enhancing tumor core, WT - whole tumor, TC - tumor core.}
	\label{tab:valid}
	\begin{tabular}{l|c|c|c|c|c|c}
		\hline
		& \multicolumn{3}{c|}{Dice} & \multicolumn{3}{c}{Hausdorff (mm)}  \\ \hline
		Validation dataset & ET & WT & TC & ET & WT & TC \\ \hline
		Single Model & 0.8145 & 0.9042 & 0.8596 & 3.8048 & 4.4834 & 8.2777 \\
		Single Model + TTA & 0.8173 & 0.9068 & 0.8602 & 3.8241 & 4.4117 & 6.8413 \\
		Ensemble of 10 models  & 0.8233 & 0.9100 & 0.8668 & 3.9257 & 4.5160 & 6.8545 \\
		\hline
	\end{tabular}
\end{table}

\begin{table}
	\centering
	%\small
	\caption{BraTS 2018 testing dataset results. Mean Dice and Hausdorff measurements of the proposed segmentation method.  EN - enhancing tumor core, WT - whole tumor, TC - tumor core.}
	\label{tab:testresults}
	\begin{tabular}{l|c|c|c|c|c|c}
		\hline
		& \multicolumn{3}{c|}{Dice} & \multicolumn{3}{c}{Hausdorff (mm)}  \\ \hline
		Testing dataset  & ET & WT & TC & ET & WT & TC \\ \hline
		Ensemble of 10 models & 0.7664 & 0.8839 & 0.8154 & 3.7731 & 5.9044 & 4.8091 \\

		\hline
	\end{tabular}

\end{table}

Time-wise, each training epoch (285 cases) on a single GPU (NVIDIA Tesla V100 32GB) takes ~9min. Training the model for 300 epochs takes ~2 days. We've also trained the model on NVIDIA DGX-1 server (that includes 8 V100 GPUs  interconnected with NVLink); this allowed to train the model in ~6 hours (~7.8x speed up). The inference time is 0.4 sec for a single model on a single V100 GPU.

\section{Discussion and Conclusion}
 \label{sec:conclusion}
 
In this work, we described a semantic segmentation network for brain tumor segmentation from multimodal 3D MRIs, which won the BraTS 2018 challenge. 
%among more than 390 participating teams.  
While experimenting with network architectures, we have tried several alternative approaches. For instance, we have tried a larger batch size of 8  to be able to use BatchNorm (and take advantage of batch statistics), however due to the GPU memory limits this modification required to use a smaller image crop size, and resulted in worse performance. We have also experimented with more sophisticated data augmentation techniques, including random histogram matching, affine image transforms, and random image filtering, which did not demonstrate any additional improvements.  We have tried several data post-processing techniques to fine tune the segmentation predictions with CRF~\cite{Kamnitsas16}, but did not find it beneficial (it helped for some images, but made some other image segmentation results worse).  Increasing the network depth further did not improve the performance, but increasing the network width (the number of features/filters) consistently improved the results. Using the NVIDIA Volta V100 32GB GPU we were able to double the number of features compared to V100 16GB version. Finally, the additional VAE branch helped to regularize the shared encoder (in presence of limited data), which not only improved the performance, but helped to consistently achieve good training accuracy for any random initialization.  Our BraTS 2018 testing dataset results are 0.7664, 0.8839 and 0.8154 average dice for enhanced tumor core, whole tumor and tumor core, respectively.

%
% BibTeX users should specify bibliography style 'splncs04'.
% References will then be sorted and formatted in the correct style.
%
\bibliographystyle{splncs04}
\bibliography{am_brats2018}

\end{document}